\documentclass[runningheads]{llncs}
\usepackage{graphicx}
\usepackage{hyperref}

\usepackage{tikz}
\usepackage{comment}
\usepackage{amsmath,amssymb} 
\usepackage{color}
\usepackage{booktabs}
\usepackage{float}

\usepackage[accsupp]{axessibility}  

\usepackage{xcolor}

\begin{document}
\pagestyle{headings}
\mainmatter
\def\ECCVSubNumber{322}  

\title{Self-Supervised 3D Human Pose Estimation in Static Video Via Neural Rendering} 

\titlerunning{3D Pose Via Neural Rendering}
%
\author{Luca Schmidtke$^{1,2}$ \and
Benjamin Hou$^{1}$ \and
Athanasios Vlontzos$^{1}$ \and \\
Bernhard Kainz$^{1,2}$}
\authorrunning{L. Schmidtke et al.}
%

\institute{$^{1}$Imperial College London, UK \\
$^{2}$Friedrich-Alexander-Universit\"at Erlangen-N\"urnberg, DE}

\maketitle

\begin{abstract}
Inferring 3D human pose from 2D images is a challenging and long-standing problem in the field of computer vision with many applications including motion capture, virtual reality, surveillance or gait analysis for sports and medicine. We present preliminary results for a method to estimate 3D pose from 2D video containing a single person and a static background without the need for any manual landmark annotations. We achieve this by formulating a simple yet effective self-supervision task: our model is required to reconstruct a random frame of a video given a frame from another timepoint and a rendered image of a transformed human shape template. Crucially for optimisation, our ray casting based rendering pipeline is fully differentiable, enabling end to end training solely based on the reconstruction task.

\end{abstract}

\section{Introduction}
Inferring 3D properties of our world from 2D images is an intriguing open problem in computer vision, even more so when no direct supervision is provided in the form of labels. Although this problem is inherently ill-posed, humans are able to derive accurate depth estimates, even when their vision is impaired, from motion cues and semantic prior knowledge about the perceived world around them. This is especially true for human pose estimation. Self-supervised learning has proven to be an effective technique to utilise large amounts of unlabelled video and image sources. On a more fundamental note, self-supervised learning is hypothesised to be an essential component in the emergence of intelligence and cognition. Moreover, self-supervised approaches allow for more flexibility in domains such as the medical sector where labels are often hard to come by. In this paper we focus on self-supervised 3D pose estimation from monocular video, a key element of a wide range of applications including motion capture, visual surveillance or gait analysis.

Inspired by previous work, we model pose as a factor of variation throughout different frames of a video of a single person and a static background.
More formally, self-supervision is provided by formulating a conditional image reconstruction task: given a pose input different from the current image, what would that image look like if we condition it on the given pose? Differently from previous work, we choose to represent pose as a 3D template consisting of connected parts which we transform and project to two-dimensional image space, thereby inferring 3D pose from monocular images without explicit supervision.

More specifically, our method builds upon the recent emergence and success of combining deep neural networks with an explicit 3D to 2D image formation process through fully differentiable rendering pipelines. This inverse-graphics approach follows the analysis by synthesis principle of generative models in a broader context: We hope to extract information about the 3D properties of objects in our world by trying to recreate their perceived appearance on 2D images. Popular rendering techniques rely on different representations including meshes and polygons, point clouds or implicit surfaces. In our work we make use of volume rendering with a simple occupancy function or density combined with a texture field that assign an occupancy between $[0, 1]$ and RGB colour value $c \in \mathbb{R}^3$ for every point defined on a regular 3D grid.\\

\section{Related Work}
\noindent\textbf{Monocular 3D Human Pose Estimation}
Human pose estimation in general is a long standing problem in computer vision with an associated large body of work and substantial improvements since the advent of deep-learning based approaches. Inferring 3D pose from monocular images however remains a challenging problem tackled by making use of additional cues in the image or video such as motion or multiple views from synchronised cameras or introducing prior knowledge about the hierarchical part based structure of the human body.

\noindent\textbf{Lifting from 2D to 3D}
Many works break down the problem into first estimating 2D pose and subsequently estimate 3D pose either directly \cite{JMartinez:ICCV:2017}, by leveraging self-supervision through transformation and reprojection \cite{Li_Li_Jiang_Zhang_Huang_Xu_2020} or a kd-tree to find corresponding pairs of detected 2D pose and stored 3D pose \cite{Iqbal2018ADA}. 

\noindent\textbf{Motion Cues From Video}
Videos provide a rich source of additional temporal information that can be exploited to limit the solution space. \cite{DBLP:journals/corr/LinLLWC17}, \cite{DBLP:journals/ijcv/KatirciogluTSLF18}, \cite{DBLP:conf/eccv/HossainL18} and \cite{vibe} use recurrent architectures in the form of LSTMs or GRUs to incorporate temporal context while \cite{pavllo:videopose3D:2019} employ temporal convolutions and a reprojection objective.

\noindent\textbf{Multiple Views}
Other approaches incorporate images from multiple, synchronised cameras to alleviate the ill-posedness of the problem. \cite{Pavlakos2017CoarsetoFineVP}, \cite{Tom2018RethinkingPI} and  \cite{Qiu2019CrossVF} fuse multiple 2D heatmaps while \cite{Rhodin2018UnsupervisedGR} and\cite{Rhodin2018LearningM3} utilize multi-view consistency as a form of additional supervision in the objective function.

\noindent\textbf{Human Body Prior}
Using non-paremetric belief propagation, \cite{black_mixture_experts} estimate the 2D pose of loosely-linked human body parts from image features and use a mixture of experts to estimate a conditional distribution of 3D poses. Many more recent approaches rely on features extracted from convolutional neural networks \cite{cnn}. Many works such as \cite{mesh_recovery}, \cite{spin} and \cite{vibe} make use of SMPL \cite{SMPL}, a differentiable generative model that produces a 3D human mesh based on disentangled shape and pose parameters. \cite{Xu_2020_CVPR} leverage kinematic constraints to improve their predictions while \cite{Kundu2020SelfSupervised3H} leverage a forward kinematics formulation in combination with the transformation of a 2D part-based template to formulate self-supervision in form of image reconstruction similar in some ways to our approach.

\noindent\textbf{Human Neural Rendering}
Recently, neural rendering approaches, \textit{ie.} fully differentiable rendering pipelines, have gained a lot of attention. Volume rendering techniques [\cite{Lombardi2019NeuralV}, \cite{Mildenhall2020NeRFRS}] have been demonstrated to be powerful tools to infer 3D properties of objects from 2D images when used in combination with neural networks. The end-to-end differentiability offers the intriguing opportunity to directly leverage pixel-wise reconstruction losses as a strong self-supervision signal.
This has sparked a number of very recent works estimating human 3D shape and pose via neural radiance fields \cite{Mildenhall2020NeRFRS} [\cite{Kwon2021NeuralHP}, \cite{Su2021ANeRFAN}] or signed-distance function based rendering \cite{Jiang2022SelfReconSR}.

\section{Method}
\begin{figure*}[t]
\begin{center}

  \includegraphics[width=0.9\linewidth]{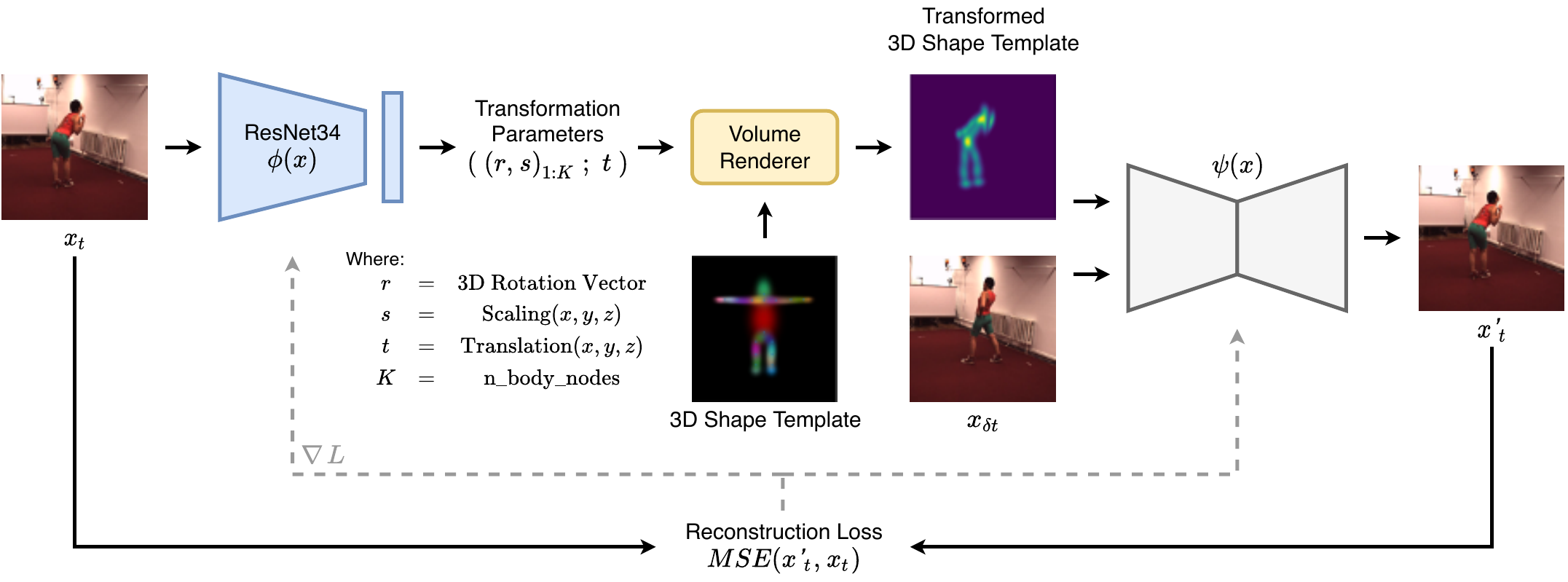}
  \caption{Our method -- left to right-- An input frame $x_t$ passes through the pose extractor encoder $\phi$ and produces the transformation parameters for each skeletal node of the shape template $\textbf{T}$. The transformed template is then rendered and concatenated with a random frame of the same sequence, $x_{\delta t}$, and is passed into an auto-encoder that's tasked to reconstruct the original frame.  
  \label{fulltrainmethod}}
\end{center}
\end{figure*}

Our approach relies on self-supervision through image reconstruction conditioned on a transformed and rendered shape template. The images are sampled from a video containing a single person moving in front of a static background. More formally, the goal is to reconstruct a number of frames $(\mathbf{x}_{t_1}, \mathbf{x}_{t_n})$ from random time points $t_1, ..., t_n$ in a video with access to \textit{one} frame $\mathbf{x}_{t_k}$, again sampled randomly, and rendered images of transformed templates  $\mathbf{T_1}, ..., \mathbf{T_n}$. 

Our method can be viewed in two distinct steps; regression of template transformation parameters and image reconstruction, where both steps are parameterized using deep convolutional neural networks.
An encoder network $\phi$ regresses rotation, translation and scale parameters from frame $\mathbf{x}_t$ in order to transform each skeletal node of a 3D shape template, $\mathbf{T}$. The generator network $\psi$ takes as input; (a) a frame $\mathbf{x}_{\delta t}$, from a different time instance of the same sequence where the same person assumes a different pose, and (b) a rendered image of the transformed 3D template while being tasked to reconstruct frame $\mathbf{x}_{t}$. 

The encoder $\phi$ consists of a convolutional neural network for feature extraction followed by a number of linear layers and a reshape operation. The generative network $\psi$ resembles a typical convolutional encoder-decoder structure utilised for image translation, where feature maps are subsequently downsampled via strided convolutions and the number of features increases. For the decoder we utilise bilinear upsampling and spatially adaptive instance normalisation (SPADE) \cite{spade} to facilitate semantic inpainting of the rendered template image.

\subsection{Template and Volume Rendering}
\subsubsection{Shape Template:} A shape template, $\mathbf{T}$, consists of $K$ Gaussian ellipsoids that are arranged in the shape of a human. Each skeletal node, denoted as $\mathbf{T}_k$, is defined on a regular volumetric grid and represents a single body part. All ellipsoids are parameterized by their mean $\mathbf{\mu}_k$ and co-variance $\mathbf{\Sigma}_k$. On the volumetric grid wee define two functions: a scalar field $f: \mathbb{R}^3 \rightarrow [0, 1]$ that assigns a value to each point ($x, y, z)$ on the grid --- in the volume rendering literature it is commonly referred to as the occupancy function, and a vector field $c: \mathbb{R}^3 \rightarrow \mathcal{C} \subset \mathbb{R}^3$ specifying the RGB-colour for each point, commonly referred to as the colouring function.

\subsubsection{Raycasting and Emission Absorption Function:}
We make use of an existing implementation of the raycasting algorithm shipped with the PyTorch3D package \cite{ravi2020pytorch3d} to render the template image. Given a camera location $\mathbf{r}_0 \in \mathbb{R}^3$, rays are ``emitted'' from $\mathbf{r}_0$ that pass through each pixel $\mathbf{u}_i \in \mathbb{R}^3$ lying on a 2D view plane $\mathcal{S}$ by sampling uniformly spaced points along each ray starting from the intersecting pixel: 
\begin{equation}
    \mathbf{p}_j = \mathbf{u}_i + j \delta s,
\end{equation}
where $j$ is the step and $\delta s$ the step size that depends on the maximum depth and number of points along each ray.

The colour value at each pixel location $\mathbf{u}_i$ is then determined by a weighted sum of all colour values of the points sampled along the ray:
\begin{equation}
    \mathbf{c}_i = \sum_{j=0}^{J} w_j \mathbf{c}_j
\end{equation}
The weights $w_j$ are computed by multiplying the occupancy function $f(x)$ with the transmission function T(x) evaluated at each point $\mathbf{p}$ along the ray:
\begin{equation}
    w_j = f(\mathbf{p}_j) \cdot T(\mathbf{p}_j),
\end{equation}
where $T(\mathbf{x})$ can be interpreted as the probability that a given ray is not terminated, \emph{i.e.} fully absorbed, at a given point $\mathbf{x}$ and is computed as the cumulative product of the complement of the occupancy function of all $k$ points up until $\mathbf{p}_j$:
\begin{equation}
    T(\mathbf{x}) = \prod_k (1-f(\mathbf{x}_k))
\end{equation}
Repeating this for all pixels in the view plane results in a 2D projection of our 3D object representing the rendered image $\mathbf{f}_r \in \mathbb{R}^{3 \times h \times w}$.

\subsection{Pose regression and shape transformation}
In order to estimate the skeletal pose of a given frame, we use the encoder network, $\phi: \mathbb{R}^{3\times h \times w} \rightarrow \mathbb{R}^{3K + 3}$ based on the ResNet-34 architecture~\cite{He2016DeepRL}.

The encoder maps a color input image of size $h \times w$ to $K$ rotation and scale vectors, $(\mathbf{r},\mathbf{s})_{1:K} \in \mathbb{R}^3$, and a single global translation vector, $\mathbf{t} \in \mathbb{R}^3$ for the camera. $K$ denotes the number of transformable parts in the template. Here, rotation is parameterised via axis-angle representation, and is subsequently converted to 3D transformation matrices using the Rodrigues' rotation formula. Combined with the scaling parameter for each axis, the resulting matrix defines the affine mapping, excluding the sheer component, for spatial transformation of each skeletal node.

After construction of the 3D transformation matrix, each Gaussian ellipsoid of the template $\mathbf{T}$, with occupancy $f_k(\mathbf{x})$ and colour field $c_k(\mathbf{x})$,  gets transformed according to the regressed parameters. Finally, utilising the aforementioned ray-tracing method we render an image based upon our transformed template by summing together all transformed occupancy and colour fields and clipping to a maximum value of 1:
\begin{align} 
    &\mathbf{\Tilde{T}}_k = \Omega_k(\mathbf{T}_k) \\
    &\mathbf{f}_r = \mathcal{R}\big(\sum_k f_k(\mathbf{x}), \sum c_k(\mathbf{x})\big),\label{render_eq}
\end{align}
,where $\mathcal{R}$ denotes the rendering operation.

\subsection{Kinematic chain}
\begin{figure}[]
    \begin{center}
    \includegraphics[width=0.8\linewidth]{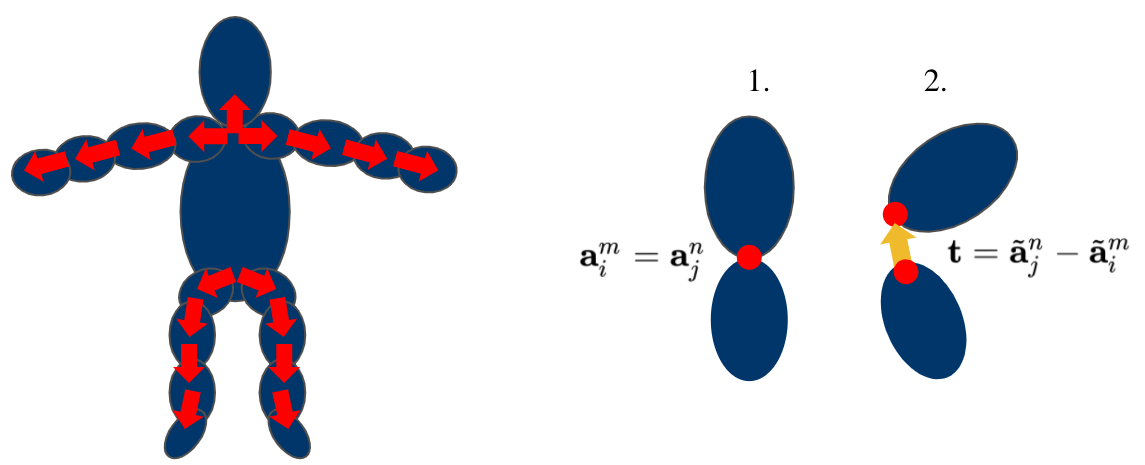}
    \end{center}
    \caption{Illustration of the kinematic chain. Red circles denote anchor-points. Following a transformation of the upper part, the translation $\mathbf{t} = \tilde{\mathbf{a}}_j^n - \tilde{\mathbf{a}}_i^m $ is applied to enforce continuity }
    \label{fig:kinematic_chain}
\end{figure}

Instead of relying on an additional loss to enforce connectivity between body parts as in \cite{LSchmidtke2021}, we define a kinematic chain along which each body part is reconnected to its parent via a translation after rotation and scale have been applied.

\noindent Given a parent and child body part with indices $n$ and $m$ respectively, we define anchor-points $\mathbf{a}_i^n$ and $\mathbf{a}_j^m \in \mathbb{R}^3$  on each part representing the area of overlap in the non-transformed template (see Figure \ref{fig:kinematic_chain} right). If body part $m$ is being transformed, the position of the anchor point changes: $\Tilde{\mathbf{a}}_j^m = H \mathbf{a}_j^m$, where $H \in \mathbb{R}^{3 \times 3}$ specifies a transformation matrix. To ensure continuous connectivity, we apply the transformation for the child body part in an analogous way, it is reconnected with the parent node by applying translation $\mathbf{t} = \tilde{\mathbf{a}}_j^n - \tilde{\mathbf{a}}_i^m $. We first transform the core, and then proceed with all other parts in an iterative fashion along the kinematic chain as depicted in Figure \ref{fig:kinematic_chain} left.

\subsection{Loss function}
The loss function is a sum of individual components: the reconstruction loss as the pixel-wise $l^2$-norm between the decoder output and the original image and a boundary loss of the form
\begin{equation}
    \mathcal{L}_{bx}^i=
    \begin{cases}
      \vert \hat{a}_{x,i}\vert, & \text{if}\ \vert\hat{a}_{x,i}\vert > 1 \\
      0, & \text{otherwise}
    \end{cases}
   \quad \quad \mathcal{L}_{bx} = \sum_{i} \mathcal{L}_{bx}^i
   \quad \quad \mathcal{L}_b = \mathcal{L}_{bx} + \mathcal{L}_{by}
\end{equation}
,where $\hat{a}_{x,i}$ is the x-component of a projected and transformed anchor point. Note that we normalise image coordinates to $(-1, 1)$.
\\
We also regularise the pose regression via the  $l^2$-norm of the rotation vector $\mathbf{r}$ and decay this term linearly to 0 after 500 iterations.
Overall, our objective function is:
\begin{equation}
    \mathcal{L} = \mathcal{L}_{\text{recon}} + \mathcal{L}_b + \alpha * \sum_K \vert \vert \mathbf{r}_k \vert \vert_2, \quad \alpha = \text{min}(1-0.02*\text{iter})
\end{equation}

\section{Experiments}
We train and evaluate our model on Human 3.6M \cite{h36m_pami}, a motion-capture dataset including 11 actors performing various activities while being filmed by four different cameras in a studio setting. Following \cite{Jakab2020SelfSupervisedLO} and \cite{LSchmidtke2021}, we train on subjects 1, 5, 6, 7, test on 9 and 11 and restrict activities to mostly upright poses, resulting in roughly 700,000 images for training. We sample video frames in pairs containing the same person in different poses, but with the same background utilising bounding boxes derived from the masks and utilise the Adam optimizer \cite{Kingma2015AdamAM}.

\section{Results}
\begin{figure*}[t]
\begin{center}

 \includegraphics[width=\linewidth]{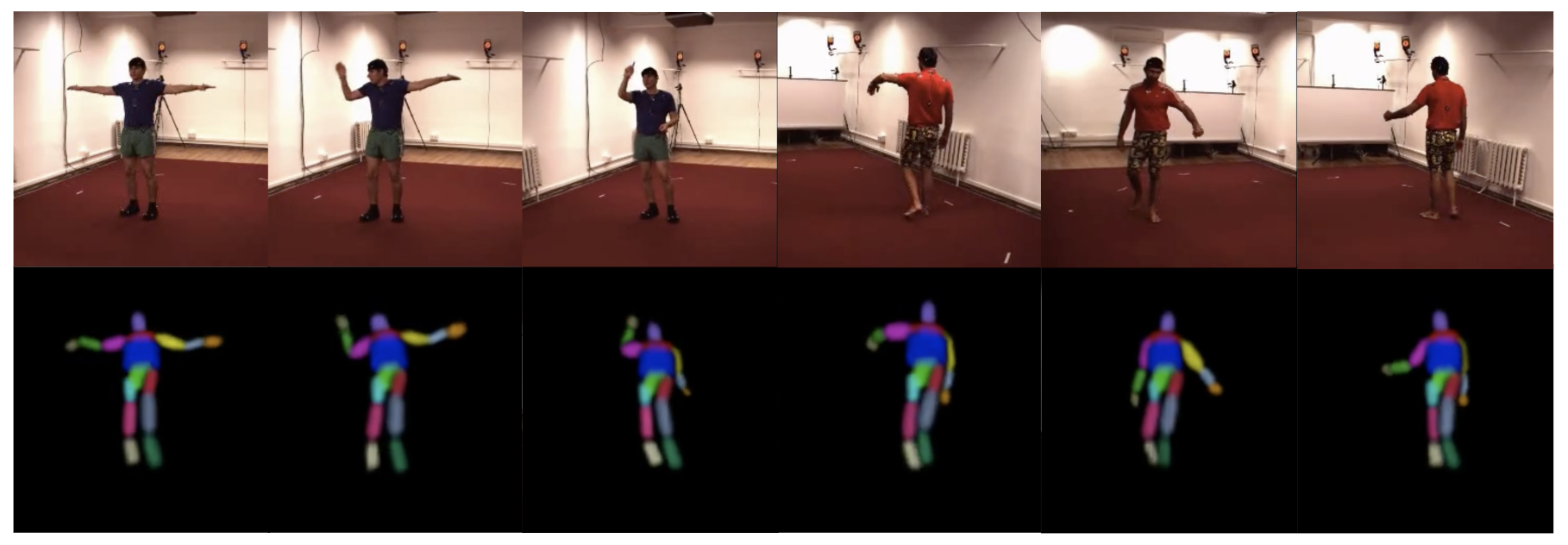}
 \caption{Results on the two evaluation subjects. Top row: input image. Bottom row: predicted pose in the form of a transformed and rendered 3d shape template.}
 \label{figure:results}
\end{center}
\end{figure*}

We restrict our evaluation to qualitative results in figure \ref{figure:results}. These demonstrate that the concept of self-supervision through conditional image translation can be extended to 3D pose estimation. However, there are several issues that still need to be solved: The model is currently not able to distinguish left and right, as can be observed in figure \ref{figure:results} (fourth image from the right), where the subject is facing away from the camera, but the template remains in the front-facing configuration.
The model also mostly generates limbs facing away from the camera (third image from the right). We hypothesise that due to depth ambiguity in 2D and limitations in pose variety due to the restricted sampling the decoder can perfectly reconstruct the image despite the wrong orientation of the limb. 

\section{Conclusion}
We presented preliminary results for a method to estimate human pose in 3d from monocular images without relying on any landmark labels. Despite issues with depth ambiguity the qualitative results are encouraging and demonstrate the feasibility of combining differentiable rendering techniques and self-supervision. A straightforward improvement would be weak supervision in the form a small labelled dataset. Replacing the image translation task with a purely generative approach with separate fore- and background similarly to \cite{Yang2022LearningFS} might prove to be very successful in extending the approach to non-static backgrounds as well.
\\ \\
\noindent\textbf{Acknowledgements:}
 supported by EPSRC EP/S013687/1.

\bibliographystyle{splncs04}
\bibliography{egbib}
\end{document}